\documentclass[10pt,twocolumn,letterpaper]{article}

\usepackage{cvpr}
\usepackage{times}
\usepackage{epsfig}
\usepackage{graphicx}
\usepackage{amsmath}
\usepackage{amssymb}
\usepackage{xcolor}
\usepackage{subcaption}

\usepackage{array}

\usepackage[breaklinks=true,bookmarks=false]{hyperref}

\cvprfinalcopy 

\def\cvprPaperID{****} 

\ifcvprfinal\fi
\begin{document}

\definecolor{brown}{rgb}{0.5,0.4,0.25}
\definecolor{darkgreen}{rgb}{0,0.4,0}
\newcommand{\mi}[1]{\textcolor{darkgreen}{{ MI: #1}}} 
\newcommand{\ar}[1]{\textcolor{brown}{{ AR: #1}}}  
\newcommand{\todo}[1]{\textcolor{blue}{{ TODO: #1}}} 
\newcommand{\att}[1]{\textcolor{red}{{#1}}} 
\renewcommand{\thefootnote}{\fnsymbol{footnote}}

\title{Interactive Video Object Segmentation in the Wild}

\author{Arnaud B\'enard\footnotemark[1]\\
gifs.com\\
{\tt\small arnaud@gifs.com}
\and
Michael Gygli\footnotemark[1] \\
gifs.com\\
{\tt\small michael@gifs.com}
}

\maketitle
\thispagestyle{empty}

\footnotetext[1]{Authors contributed equally}

\begin{abstract}
In this paper we present our system for human-in-the-loop video object segmentation.
The backbone of our system is a method for one-shot video object segmentation~\cite{caelles2017one}.
While fast, this method requires an accurate pixel-level segmentation of one (or several) frames as input. As manually annotating such a segmentation is impractical, we propose a deep interactive image segmentation method, that can accurately segment objects with only a handful of clicks.
On the GrabCut dataset, our method obtains 90\% IOU with just 3.8 clicks on average, setting the new state of the art. Furthermore, as our method iteratively refines an initial segmentation, it can effectively correct frames where the video object segmentation fails, thus allowing users to quickly obtain high quality results even on challenging sequences.
Finally, we investigate usage patterns and give insights in how many steps users take to annotate frames, what kind of corrections they provide, \etc, thus giving important insights for further improving interactive video segmentation.
\end{abstract}

\section{Introduction}
Interactive Object segmentation has been recently become popular to quickly edit photos.
Snapchat Backdrop~\cite{snapchat_backdrop}, for example, allows to change the background of portrait pictures, after outlying the foreground. In their portrait mode product, Google \cite{portrait_mode} uses object segmentation to blur out the background of a photo.
The practical usage of video object segmentation, on the other hand, has previously been limited, due to its challenges. Videos are often harder to segment due to motion blur, bad composition, occlusion, \etc. Fully automatic methods typically fail to accurately segment more challenging sequences. The other extreme, requiring user input for each frame, is impractical due to its time requirements.
Thus, most research in video object segmentation adapts a semi-supervised approach. There, a subset of one or several frames in a sequence are manually segmented and used to infer the segmentation masks of all frames in the sequence~\cite{caelles2017one,lucid_dreaming, Pont-Tuset_arXiv_2017}.

\begin{figure}[t]
	\centering
	\includegraphics[width=1\linewidth]{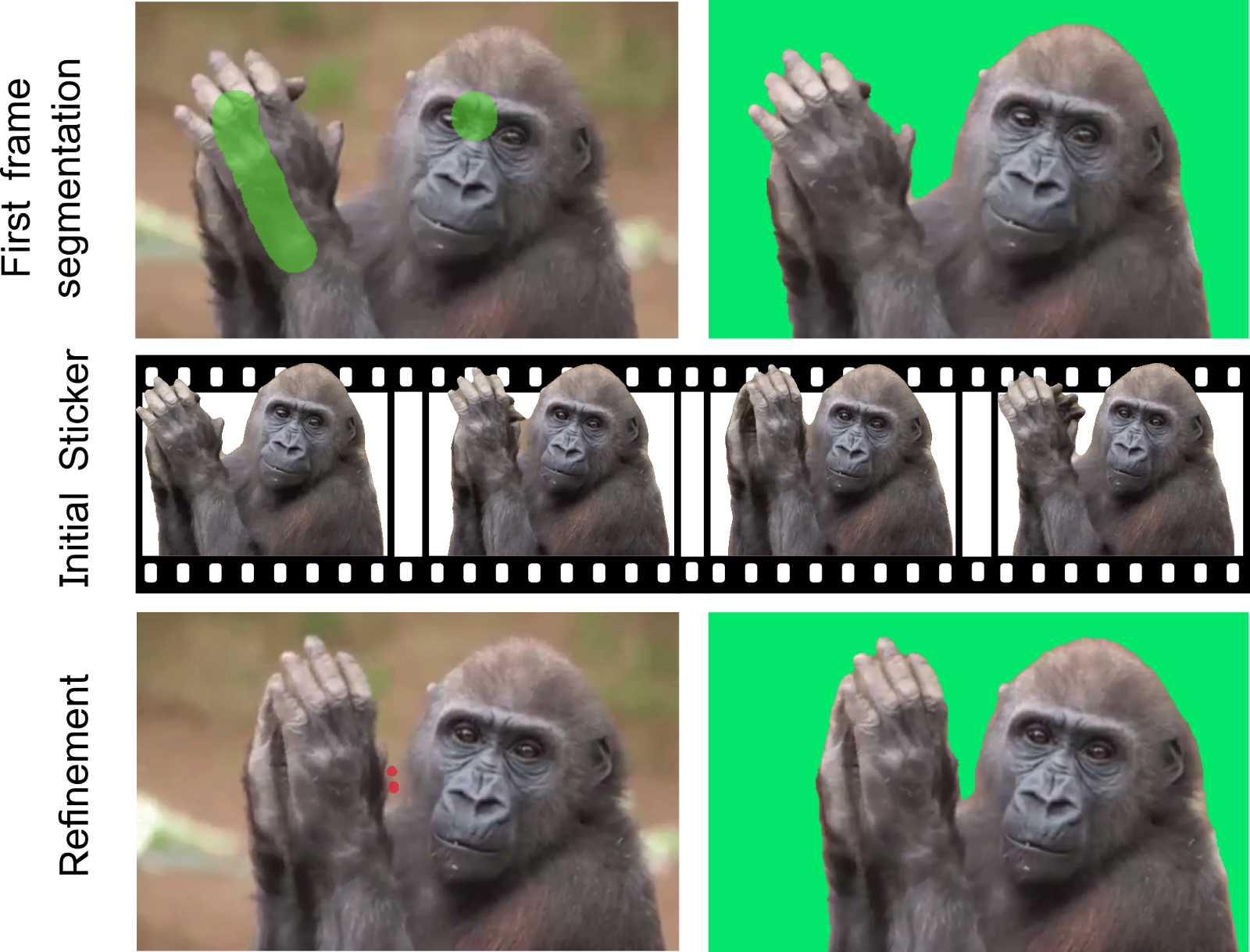}
	\caption{Overview of our method. The first frame is segmented with a handful of clicks.
	Then, the video snippet is segmented using OSVOS~\cite{caelles2017one}.
	Finally, the user can optionally refine poorly segmented frames with 1-2 clicks.}
    \label{fig:teaser}
\end{figure}
All aforementioned methods assume that the training frames are manually annotated with pixel-accurate segmentations. This prevents their practical usage, as annotating a single frame typically takes more than a minute~\cite{maninis2017deep}.
In this paper, we build upon the semi-supervised approach of~\cite{caelles2017one}. But our work makes it practical by proposing a fast interactive object segmentation algorithm, which allows to segment the first frame in a few seconds (see Figure~\ref{fig:teaser}).
In our experiments, we show that using these masks, rather than costly pixel-accurate segmentations, leads to a minimal performance degradation ($-3.2\%$ IOU).

Our method for interactive segmentation is based on~\cite{xu2016deep}, which uses a deep convolutional neural network that jointly uses user clicks and RGB information to segment images.
The key idea of our approach is to use the current segmentation mask as an additional input. Thus, our method uses the user clicks to \textit{refine} an initial segmentation. As our experiments show, our model offers high image segmentation quality and -- more importantly -- allows to quickly correct segmentations obtained by another method, OSVOS~\cite{caelles2017one} in our case.
Our methodological improvements make our system usable practical and allowed us to publicly release it.
By monitoring usage, we were able to obtain insights into its usage patters, something that is important for designing future video segmentation systems. We find, for example, that there are three common types of annotation and that the quality is much lower \textit{in the wild} compared to the controlled settings that prevail in video object segmentation~\cite{Perazzi2016}.

To summarize, this work makes the following contributions:
\begin{itemize}
  \item A system for interactive video object segmentation based on~\cite{caelles2017one}.
  \item A novel method for deep interactive object segmentation that iteratively refines an initial segmentation by using user input in the form of clicks.
  \item A thorough experimental evaluation of our method and a comparison to state-of-the-art interactive segmentation methods. Our method sets a new state of the art on the GrabCut dataset~\cite{rother2004grabcut}.
  We further show that it works well for providing a fast initialization for OSVOS.
  \item An analysis of annotation patterns on our production data.
\end{itemize}

\section{Method}
\begin{figure}[t]
	\centering
\frame{\includegraphics[width=0.49\linewidth]{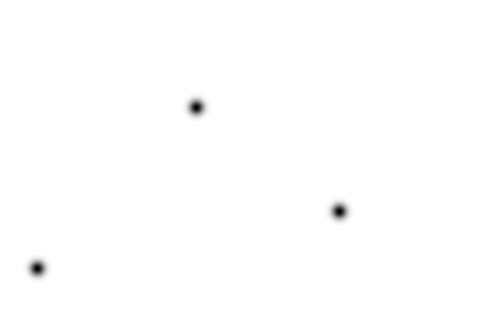}}
	\frame{\includegraphics[width=0.49\linewidth]{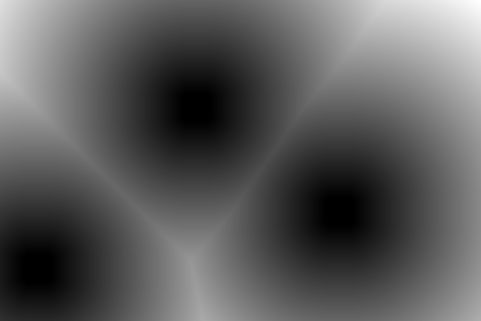}}
	\caption{Gaussians placed at click positions \vs Euclidean distance map, in this case for positive clicks.}
    \label{fig:gaussian_vs_dist}
\end{figure}
\begin{figure}[t]
	\centering
	\includegraphics[trim={0 0 0 2cm},clip,width=0.49\linewidth]{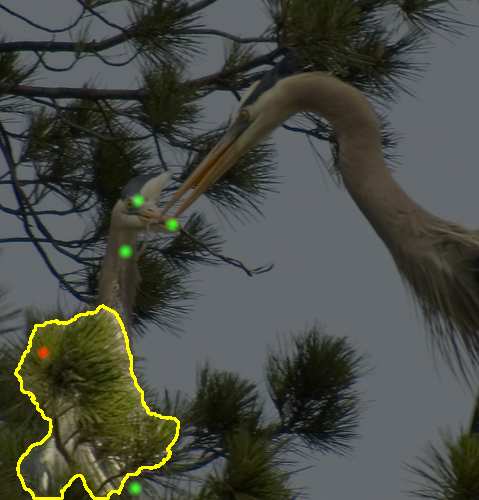}
	\includegraphics[trim={0 0 0 2cm},clip,width=0.49\linewidth]{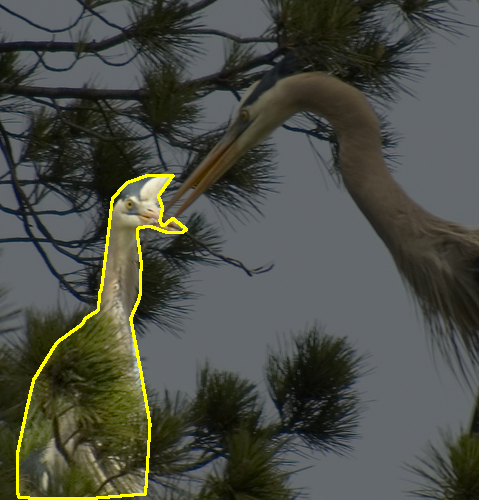}
	\caption{Sampling corrections. Left: Current prediction and the sampled correction clicks. Right: Ground Truth mask. \textcolor{green}{Positive}/\textcolor{red}{Negative} points are sampled from the error region of the initial prediction.}
    \label{fig:corr_samples}
\end{figure}
Our approach for video object segmentation consists of two main steps: The semi-automatic annotation of the first frame and the automatic propagation of this segmentation mask to the remaining frames of the video.
We will now describe these steps in turn.
Finally, as video segmentation might give unsatisfactory results on challenging sequences with small objects or a lot of variation, we show how our method trivially generalizes to the case of refining an initial video segmentation result.

\subsection{Interactive image segmentation}
\label{sec:int_seg_method}
Interactive image segmentation is always \textit{iterative} image segmentation, where a user provides input to \textit{refine} the current result. We propose an approach that can implicitly model that causality by providing the current segmentation mask as an input to the model, together with the user input.
Specifically, we propose a deep convolutional neural network, which predicts a pixel-level segmentation as a function of the current segmentation, the user inputs and the RGB image. The current segmentation is simply a binary image provided as as an extra channel.
To encode user input in the form of clicks, we create a zero-initialized channel, from which Gaussians with a small standard deviation are added or subtracted for each positive or negative click, respectively (See Figure~\ref{fig:corr_samples}).
The idea of using extra channels with user information is inspired by Xu~\etal~\cite{xu2016deep}, but they use two channels, each encoding the Euclidean distance to the closest foreground or background click, respectively. We choose Gaussians instead, as we find that they lead to improved localization accuracy, which is important for obtaining high-accuracy segmentations through iterative refinement. Furthermore, they allow to encode both types of clicks in a single channel. We provide a visual comparison of the two encodings in Figure~\ref{fig:gaussian_vs_dist}.

\subsubsection{Simulating user interactions}
As it is too expensive to manually collect user interactions, we follow Xu~\etal~\cite{xu2016deep} and simulate
interactions instead. We iteratively sample foreground clicks that are $d_{margin}=3$ away from the object boundary and $d_{step}=5$ pixels away from the previous clicks.
For sampling negatives we use the same 3 strategies proposed by~\cite{xu2016deep}:
(i) randomly sampling in a hull of $d_{hull}=40$ around the object
(ii) sampling negatives clicks from other objects in the image
and (iii) sampling negatives as in (i), but such that they are maximally distant from each other, \ie surround the object of interest.

In addition to that, we generate examples that correct an initial mask.
For this, we sample clicks as above, run them through the model that is trained without initial masks to obtain a predicted segmentation. Given these predictions, we sample $N_{corr} \in \left\{0,1,2,3,4,5,10,20\right\}$ corrections clicks. Correction clicks are points that that lie in the error region. We sample them such that they are $d_{step}$ away from each other. In Figure~\ref{fig:corr_samples} we visualize this procedure on an example. Using this kind of training examples allows the model to learn to use an initial mask, in addition to user clicks. This is what allows our model to quickly correct OSVOS results.

\subsubsection{Neural Network architecture and training}
As our network architecture we choose a ResNet-101~\cite{he2016deep} model that is adapted to segmentation via the introduction of atrous convolutions and a pyramid scene parsing module~\cite{chen2017deeplab}.
We use weights from a model that is pre-trained on ImageNet and fine-tuned on PASCAL for semantic segmentation.
The network is trained to predict the foreground probability for each pixel, via a standard cross-entropy loss.
Deep neural networks cannot encode hard constraints and thus might not assign a probability of 1 for a pixel that the user labelled as foreground (and equivalently for background labelled pixels). Thus, we further clamp these pixels to be foreground/background, as in~\cite{xu2016deep}.

\subsubsection{CRF Post-processing}
Dominant CNN architectures, including the one we use, contain max-pooling and down-sampling layers. Thus, predictions are made at an lower resolution (8x in our case) and thus often poorly localized. To improve localization we propose to refine the initial predictions with a fully connected Conditional Random Field (CRF)~\cite{krahenbuhl2011efficient}:

\begin{equation}
E(\mathbf{x}) = \sum_i \theta_i(x_i) + \sum_{ij}\theta_{ij}(x_i,x_j),
\end{equation}
with unary potential $\theta_i(x_i)=-\log P(x_i)$, where $P(x_i)$ is the foreground probability of pixel $i$, as predicted by the neural network.
For the pairwise potential $\theta_{ij}(x_i,x_j)$ we use two Gaussian potentials as in~\cite{krahenbuhl2011efficient}. One, favouring nearby pixels with similar colors to take the same label and the other, favouring smoothness,~\ie encouraging to give neighbouring pixels the same label.
See~\cite{krahenbuhl2011efficient} for more information.

\subsection{Video segmentation}
For segmenting the full video, given the first frame, we rely on one-shot video segmentation.
There, the task is to segment the full video, given the segmentation of the first frame.
Specifically, we use OSVOS~\cite{caelles2017one}.
OSVOS uses a VGG-Network~\cite{simonyan2014very} that is fine-tuned for video object segmentation.
This network has thus learnt about objectness and which objects or things are likely to be foreground or background.
To adapt the network to a particular object in a video sequence, it is fine-tuned on the segmentation mask of the first frame, such that it can learn the object's appearance.
We chose OSVOS over more recent methods such as~\cite{khoreva2017lucid}, due to its speed. 
OSVOS can be fine-tuned for 500 epochs in about 1 minute. After fine-tuning, OSVOS takes only $\approx 100 ms$ per frame.	

\subsection{Progressive Refinement}
Our goal is to obtain high-quality video segmentation results, even on challenging sequences.
As one-shot segmentation might fail on such sequences, we need a way for the user to efficiently correct the initial result.~\cite{caelles2017one} proposes a progressive refinement, where the worst segmentation mask is manually annotated and used as an additional training example for fine-tuning OSVOS.
We follow this idea of correcting the worst segmentation mask, but speed up the process by relying on our interactive segmentation method presented in Section~\ref{sec:int_seg_method}. Importantly, as our method takes the current segmentation mask as input, it provides a fast and intuitive way to \textit{refine} an initial mask, rather than requiring to have it annotated from zero.
This allows to correct bad masks with very few clicks, as our experiments in Section~\ref{sec:exp_refine} show.

\section{Experiments}
\begin{table}[t]
\centering
\small
\setlength{\tabcolsep}{2.5pt}
\renewcommand{\arraystretch}{1.2}
\begin{tabular}{|l|l|l|}
\hline
\textbf{Method}  & \textbf{Pascal} &  \textbf{GrabCut} \\
\hline
iFCN w/o CRF  &	6.4 & 6.2 \\
\hline
iFCN  &	6.3 & 4.0  \\ 
\hline
Ours w/o init. mask & 5.5	 &  3.9  \\
\hline
Ours w/o CRF & 4.9 & 4.7	  \\
\hline
Ours   & 5.6 &	3.8  \\ 
\hline
\end{tabular}
\caption{Ablation study.}
\label{tab:results_detail}
\end{table}
\begin{figure*}[t]
	\centering
	 		\begin{subfigure}[t]{0.49\linewidth}
 		   	    \includegraphics[width=1\linewidth]{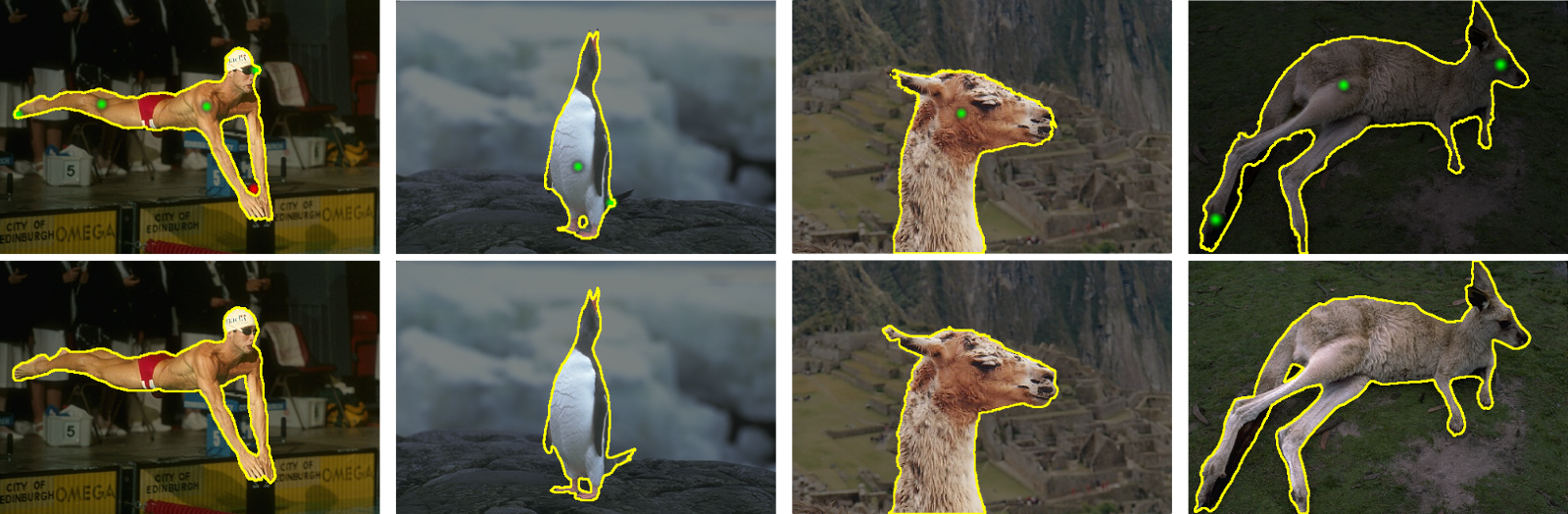} 				
				\caption{GrabCut dataset. Top: Our predictions; Bottom: Ground Truth}
   	    \end{subfigure} 
   	    \hspace{0.005\linewidth}
 		\begin{subfigure}[t]{0.49\linewidth}
 		   	    \includegraphics[width=1\linewidth]{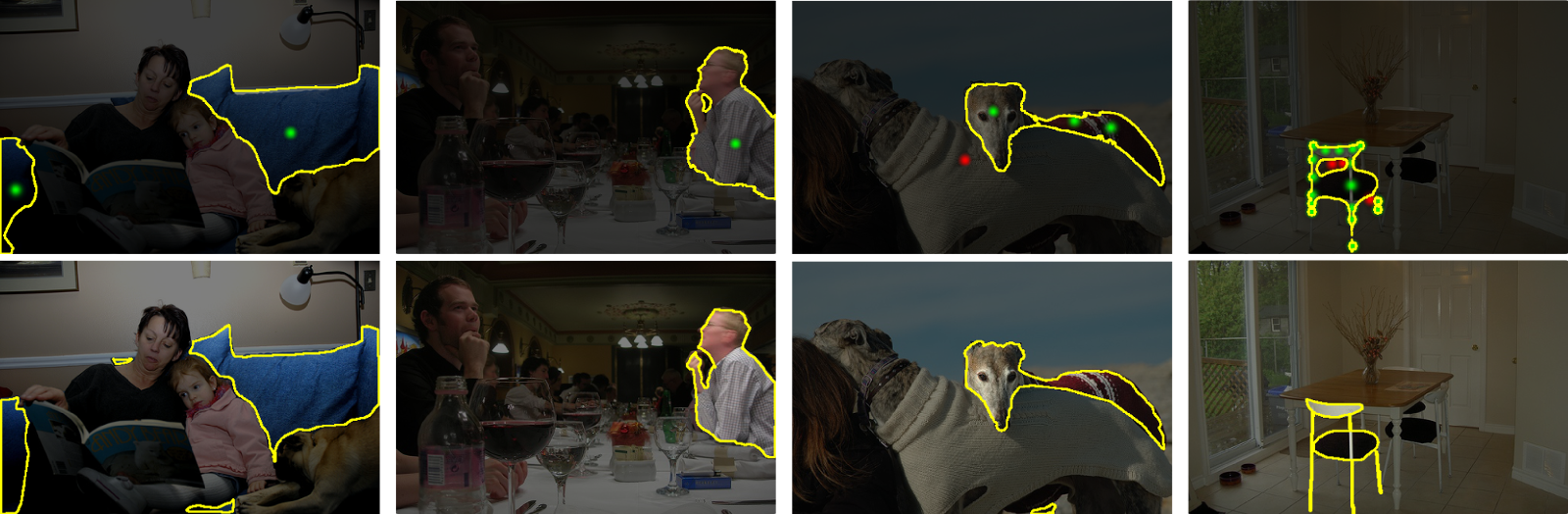} 								\caption{PASCAL dataset. Top: Our predictions; Bottom: Ground Truth}
   	    \end{subfigure}		   	    
	\caption{Results of our deep interactive object segmentation method on the GrabCut and PASCAL dataset.
	Our method can segment simple images with 1-2 clicks and also performs well when segmenting challenging, partially occluded objects.
	The rightmost image shows a failure case: It has problems segmenting thin structures such as the legs of chairs.}
    \label{fig:results_image_seg}
\end{figure*}
We evaluate our method on three publicly available datasets Pascal~\cite{everingham2011pascal}, GrabCut~\cite{rother2004grabcut} and DAVIS-2016~\cite{Perazzi2016}.
All our models are trained on 10582 images of PASCAL, augmented with the labels of SBD~\cite{hariharan2011semantic} as is common practice.

First, we evaluate our image segmentation method in Section~\ref{sec:exp_image}.
Then, in Section~\ref{sec:exp_video}, we show how video object segmentation (OSVOS~\cite{caelles2017one}) performs when initialized from a mask obtained using above method, rather than a pixel-accurate segmentation.
Finally, Section~\ref{sec:exp_refine} evaluates the ability to quickly correct an initial mask.

For evaluation we use use the mean Intersection over Union (IOU)~\cite{caelles2017one}, which is also called Jaccard index.

\subsection{Instance Segmentation in Images}
\label{sec:exp_image}
In this Section we evaluate the performance of our interactive image segmentation method of public datasets.
We compare it against state-of-the-art methods, as well as our own implementation of~\cite{xu2016deep}, using the same DeepLab network~\cite{chen2017deeplab}, CRF inference~\cite{krahenbuhl2011efficient} and trained with the augmented labels of~\cite{hariharan2011semantic}.

\subsubsection{Ablation study}
We perform an ablation study and evaluate the effect of (i) using Gaussians instead of distance maps, (ii) the CRF refinement step and (iii) using the current mask as an additional input channel.
Results are shown in Table~\ref{tab:results_detail}.

From the table we observe that using Gaussians significantly outperforms using distance maps.
This is especially pronounced when aiming for results with a high minimum IOU of 90\% and not using a CRF: iFCN w/o CRF needs $6.2$ clicks, while ours w/o CRF only needs $4.7$.
This matches our visual analysis of the results, where we find that iFCN does not always take user clicks into account,~\ie is more likely to mislabel points that were annotated by a user.
We attribute this to the linearity of the distance map, which leads to more poorly localized clicks, compared to using Gaussians (\cf Figure~\ref{fig:gaussian_vs_dist}).
The use of a CRF increases the performance on the GrabCut dataset, but decreases it on Pascal.
We believe this is due to the difficulty of the Pascal images, compared to GrabCut, thus preventing gains based on simple color statistics as used in the pairwise potentials of the CRF.
While using the CRF performs worse on PASCAL, we opt for using it in our method, as it improves boundary accuracy. Furthermore, we find that most users of our tool select large foreground objects such as people or animals, rather than small and often partially occluded objects, as they are common in PASCAL.
Thus, we believe the GrabCut dataset is closer to our real-world data.

Finally, we observe that providing the current segmentation mask as an additional channel leads to no significant performance difference. It however allows more quickly correct an initial segmentation result, as we show in Section~\ref{sec:exp_refine}, which is our main motivation for using it in our model.

\subsubsection{Comparison to the State of the Art}
\begin{table}[t]
\centering
\small
\setlength{\tabcolsep}{2.5pt}
\renewcommand{\arraystretch}{1.2}
\begin{tabular}{|l|l|l|}
\hline
\textbf{Method}  & \textbf{Pascal@85\%} &  \textbf{GrabCut@90\%}  \\
\hline
GrabCut~\cite{rother2004grabcut} &	15.06			  &	11.10			    	  	\\
iFCN~\cite{xu2016deep} 	&	6.88			  &	6.04			      	  	\\
RIS-Net~\cite{hao2017regional}    &	 5.12 			  &	5.00			  		  	\\
DEXTR \cite{maninis2017deep}    &	\textbf{4.0}  			  &	4.0			       	\\
\hline
Ours    &	5.6  			  &	\textbf{3.8}							  	\\
\hline
\end{tabular}
\caption{The mean number of clicks required to achieve a certain IOU on different datasets by various algorithms. The best results are emphasized in \textbf{bold}.}
\label{tab:stoa}
\end{table}
We compare our method against state-of-the-art methods for interactive image segmentation in Table~\ref{tab:stoa}. In particular, we compare against~\cite{xu2016deep}, on which our method is based, and the recent extension of~\cite{hao2017regional}.
We also compare against DEXTR~\cite{maninis2017deep}, which uses similar techniques, but uses extreme clicks~\cite{papadopoulos2017extreme} as input, rather than foreground/background clicks.

\noindent
\paragraph{Results.} As can be seen from the table, our method outperforms all previous methods on the GrabCut dataset. We attribute this to the use of a powerful CNN architecture (ResNet) and the use of Gaussians to encode user input. Furthermore, while~\cite{maninis2017deep} always needs at least four clicks, our method only needs 1-2 clicks on simple instances.
On the Pascal dataset, our method is outperformed by~\cite{hao2017regional,maninis2017deep}.
The objects in this dataset are small and thus challenging to segment, given their low resolution.
Indeed, both of the superior methods use techniques to crop the region of interest, thus allowing them to obtain a high resolution input even for extremely small objects.
In Figure~\ref{fig:results_image_seg} we show qualitative results. Our method often needs only 1-2 clicks to segment large objects and non-occluded objects and is also able to correctly segment objects that are partially occluded. 

\subsection{Video object segmentation}
\label{sec:exp_video}
The goal of this experiment is to measure the end-to-end accuracy of our product which outputs a full video segmentation, given a few user clicks on the first frame.
To do this, we evaluate the ability of our deep image segmentation to provide an initial mask for one-shot video object segmentation.

In the experiment, we evaluate the performance of OSVOS initialized with a first frame segmentation that is (i) the ground truth, (ii) obtained using GrabCut or (iii) obtained with our deep image segmentation.
We use a different number of clicks $N_{clicks} \in \left\{1,4,8\right\}$ as an input to the interactive segmentation methods and evaluate the performance of OSVOS on the DAVIS-2016~\cite{Perazzi2016} validation set.
For GrabCut, we provide a bounding box, in addition to the user clicks.

\begin{figure}[t]
	\centering
	\includegraphics[width=0.7\linewidth]{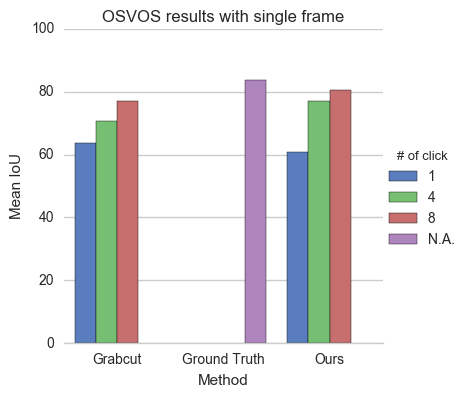}
	\caption{Comparison of OSVOS performance (mean IoU) with a single input mask generated by GrabCut, our method or taken from the ground truth.}
    \label{fig:single_frame}
\end{figure}

\noindent
\paragraph{Results.}
In Figure \ref{fig:single_frame} we compare the performance obtained by using the ground truth mask against the performance when using a mask obtained with GrabCut or our method.
Our method significantly outperforms using GrabCut, when then user provides 4 or 8 clicks. GrabCut performs better for one click, but this is because GrabCut is initialized from the ground truth bounding box, i.e. has more information. Despite this additional information provided to the GrabCut algorithm, our method performs better once the user provides more clicks.
Given 8 clicks, our method obtains a mean IOU of $80.6$, while using the ground truth has an IOU of $83.8$. Thus, using a fast interactive segmentation method, rather than a tedious and time-consuming pixel-accurate segmentation, is preferable from a user experience point of view.

We also evaluated the performance when providing input for additional frames.
We however find that adding more frames does not improve the performance significantly.

\subsection{Segmentation Refinement}
\label{sec:exp_refine}
\begin{table}[t]
\centering
\small
\setlength{\tabcolsep}{1.5pt}
\renewcommand{\arraystretch}{1.3}
\begin{tabular}{|l|l|l|l|l|} 
\hline
\textbf{Method} & \textbf{OSVOS} & \textbf{1 click} & \textbf{4 clicks} & \textbf{10 clicks} \\ \hline
GrabCut & 50.4\% & 46.6\% (-3.7)  &  53.5\% (+3.2) & 68.8\% (+18.4)\\ \hline
iFCN & 50.4\% & 55.7\% (+5.3) &  71.3\% (+20.9) & 79.9\% (+29.5)\\ \hline
Ours & 50.4\% & 63.8\% (+13.4) &  75.7\% (+25.4) & 82.2\% (+31.8)\\ \hline
\end{tabular}
\caption{Correction of bad OSVOS masks. Our method improves the masks significantly, even for few clicks, while GrabCut and iFCN need more clicks to obtain a similar improvement.}
\label{tab:results_refine}
\end{table}
\begin{figure}[t]
	\centering

			\begin{subfigure}[t]{0.33\linewidth}		
				\includegraphics[width=1\linewidth]{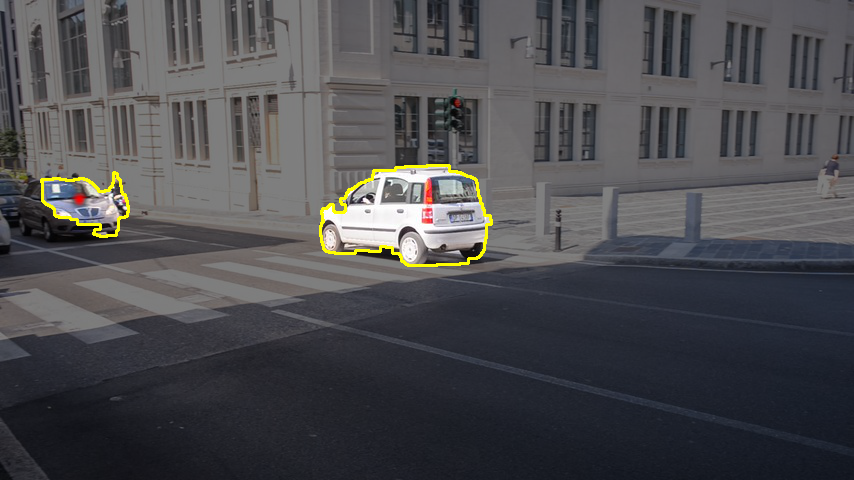} \\
						\includegraphics[width=1\linewidth]{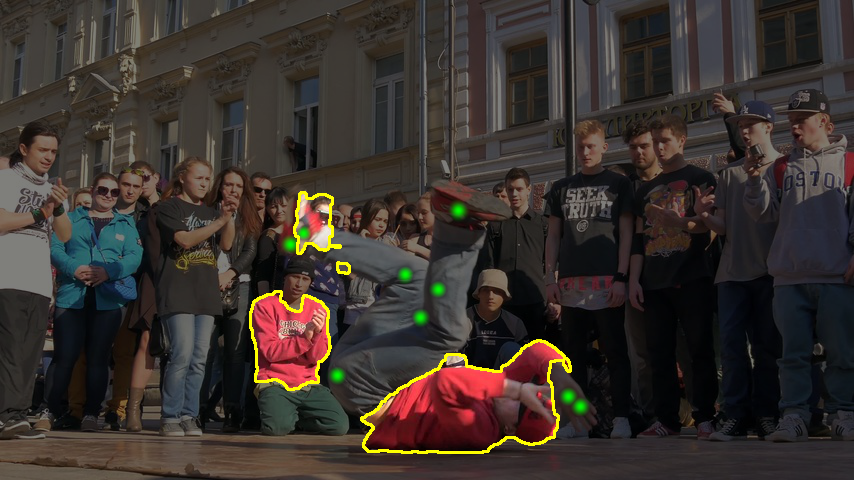}\\
		\includegraphics[width=1\linewidth]{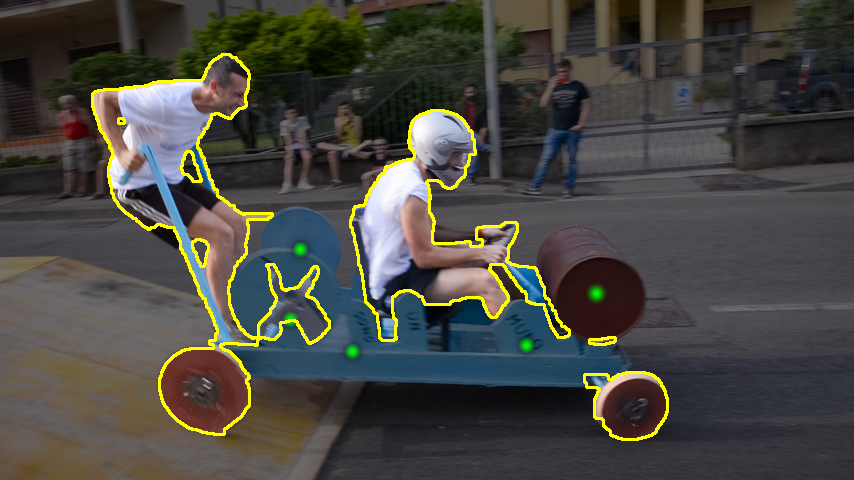}
		\caption{OSVOS result \\ and corrections}
		\end{subfigure}%
 		\begin{subfigure}[t]{0.33\linewidth}
 		   	    \includegraphics[width=1\linewidth]{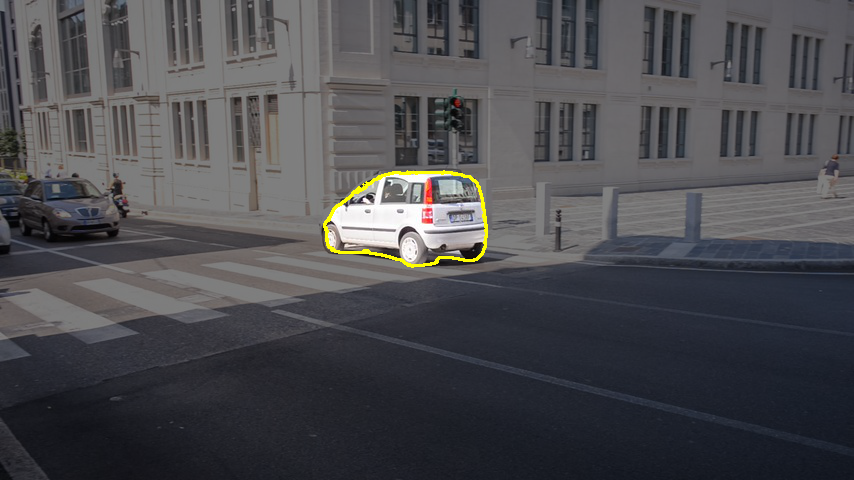}\\
 		   	       	    \includegraphics[width=1\linewidth]{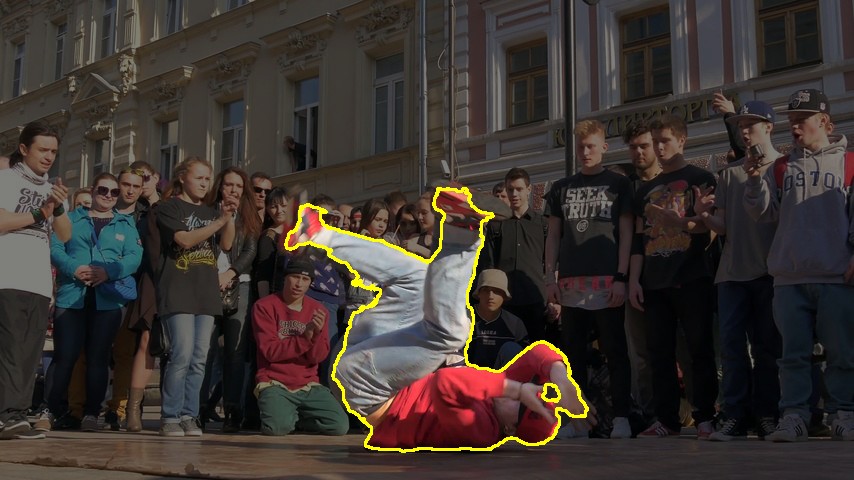}\\
   	    \includegraphics[width=1\linewidth]{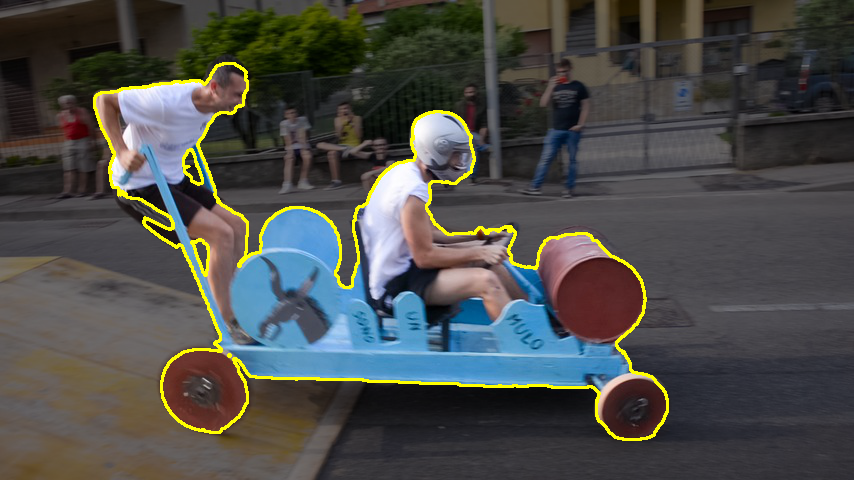}
	    		\caption{Our refined results}
   	    \end{subfigure}%
 		\begin{subfigure}[t]{0.33\linewidth}
 				\includegraphics[width=1\linewidth]{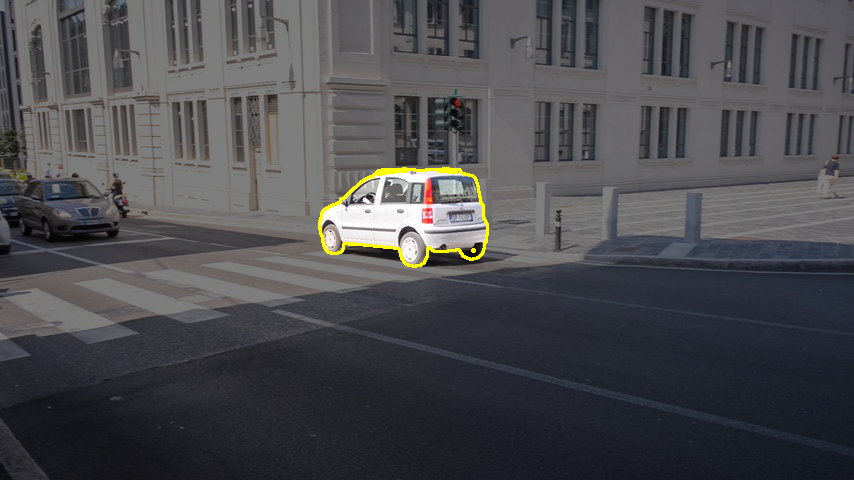} \\
 						\includegraphics[width=1\linewidth]{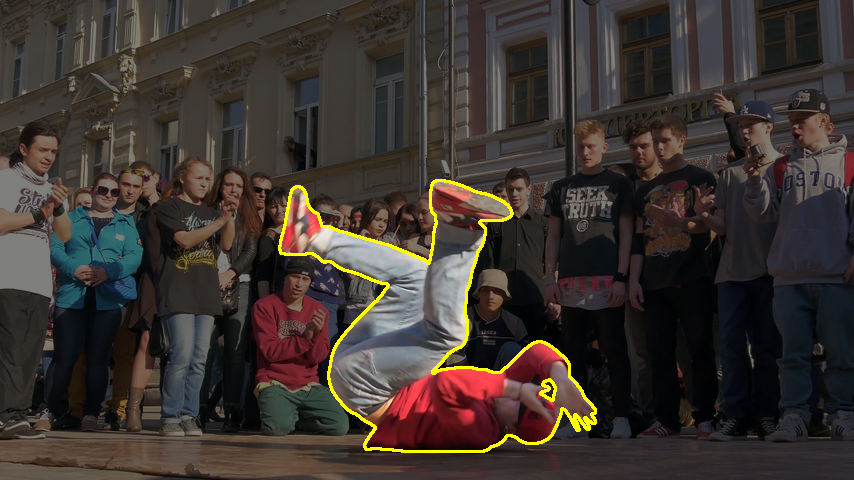} \\
		\includegraphics[width=1\linewidth]{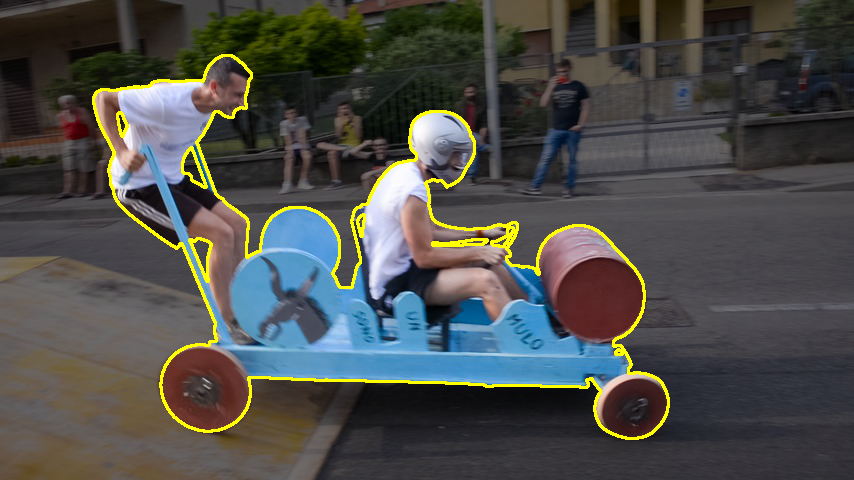}
				\caption{Ground Truth}
   	    \end{subfigure}		
	\caption{OSVOS refinement results using our interactive refinement method.}
    \label{fig:results_refine}
\end{figure}
Given that an initial result obtained with OSVOS might be of unsatisfactory quality, it is important to provide a way for making intuitive and efficient corrections.
In this section we thus evaluate the efficiency of our method in correcting an initial segmentation result.
For this, we use the DAVIS-2016~\cite{Perazzi2016} validation set. We select the worst segmentation mask per sequence, as obtained by OSVOS, and use different methods to iteratively correct the result.
Thereby we compare our method, GrabCut~\cite{rother2004grabcut} and our implementation of iFCN~\cite{xu2016deep}.

As GrabCut performs poorly without a bounding box (pixels that are certain background), we heuristically create a bounding box, which is created such that it includes all user foreground clicks and all probable foreground (the current segmentation mask). Without this trick, using GrabCut degrades the result compared to the initialization, even when using 10 correction clicks.
To initialize~\cite{xu2016deep}, we use the center of the largest foreground blob as obtained by OSVOS as a foreground click. We do not make it a hard constraint to allow the model to recover, should this initial click be incorrect.

\noindent
\paragraph{Results.} We show quantitative results in Table~\ref{tab:results_refine} and 
visual results of our method in Figure~\ref{fig:results_refine}.
As can be seen from the table, our method outperforms both baselines. The performance difference is especially prominent for few clicks: Given a single click, iFCN~\cite{xu2016deep} has only a moderate improvement of 5.3\% and GrabCut even decreases the performance. Our method, on the other hand, increases the IoU of the masks by as much as 13.4\%. Thus, this highlights the importance of initializing a method with the existing result.

\section{Analysis}
Since we released the tool for video segmentation, termed \textit{sticker editor}, our users have created hundreds of segmentations from videos and images. We call these segmentation \textit{animated stickers}. In this section, we analyze a subset (437 stickers) of our production data.

\subsection{Usage patterns}
\begin{figure}[t]
	\centering
	 		\begin{subfigure}[t]{0.32\linewidth}
 		   	    \includegraphics[width=1\linewidth]{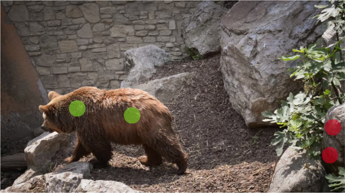} 				
				\caption{Clicks}
   	    \end{subfigure} 
 		\begin{subfigure}[t]{0.32\linewidth}
 		   	    \includegraphics[width=1\linewidth]{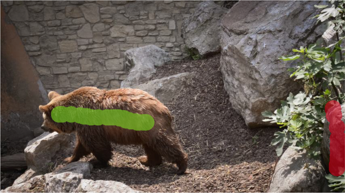} 								\caption{Strokes}
   	    \end{subfigure}		 
 		\begin{subfigure}[t]{0.32\linewidth}
 		   	    \includegraphics[width=1\linewidth]{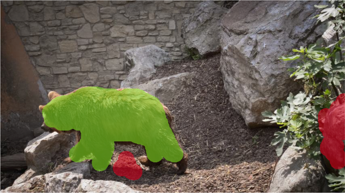} 								\caption{Highlight}
   	    \end{subfigure}		 
	\caption{Showcase of the three types of annotations we observed from analyzing our users' behavior.}
    \label{fig:annotation_patterns}
\end{figure}

The users' annotation patterns fall into three categories (as illustrated in Figure \ref{fig:annotation_patterns}): independent clicks, strokes and highlighting. Therefore, our interactive image segmentation model needs to infer accurate results with the former patterns as input.
In practice, to make our model more robust to the various kind of inputs, we thus train our model with a combination of simulated clicks and strokes. Without that, we find that our model does not perform well when a user provides strokes rather than clicks.

\begin{figure}[t]
	\centering
	\includegraphics[width=1\linewidth]{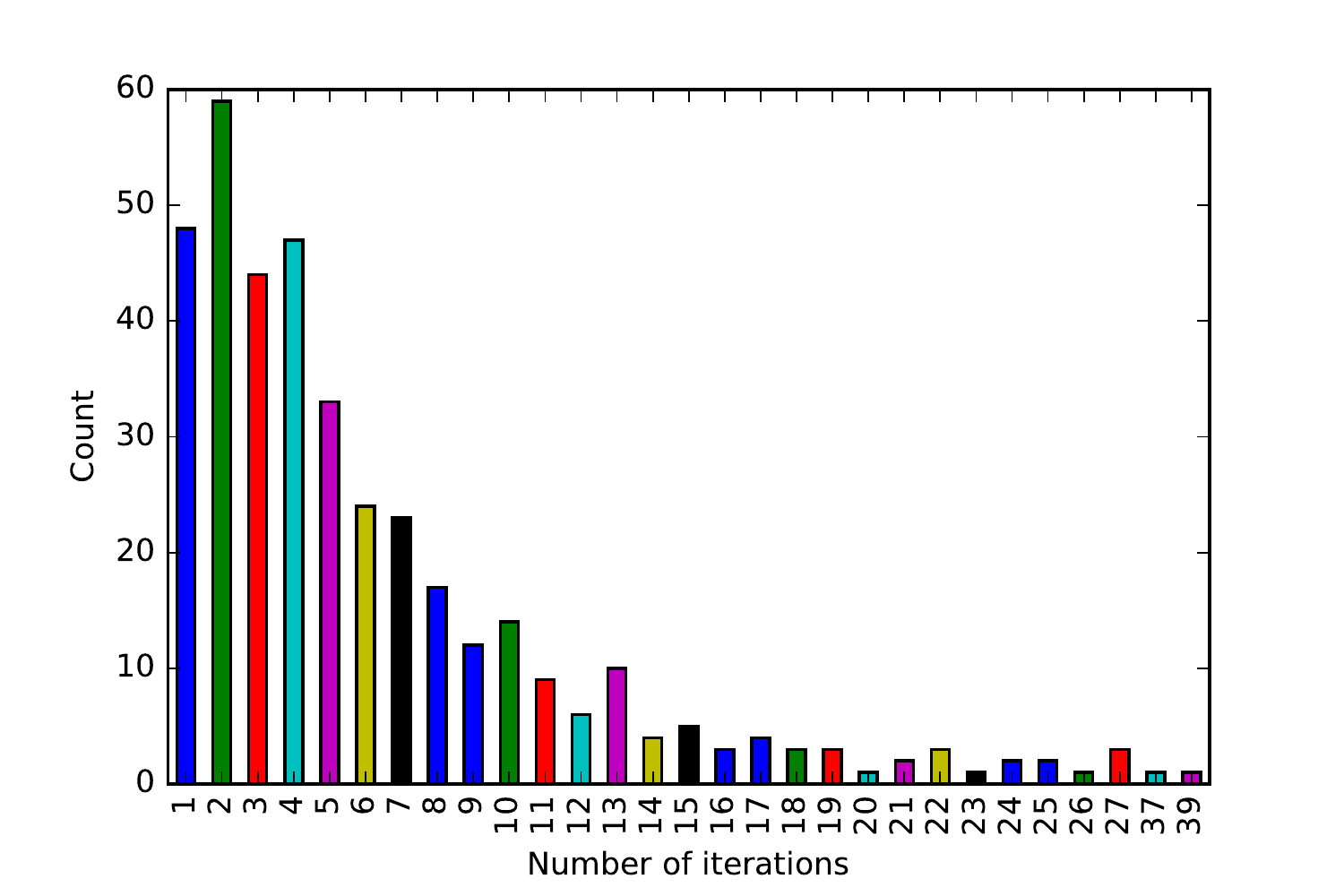}
	\caption{Histogram of the number of iterations for segmenting the first frame.}
    \label{fig:hist_iterations_0}
\end{figure}

\begin{figure}[t]
	\centering
	\includegraphics[width=0.7\linewidth]{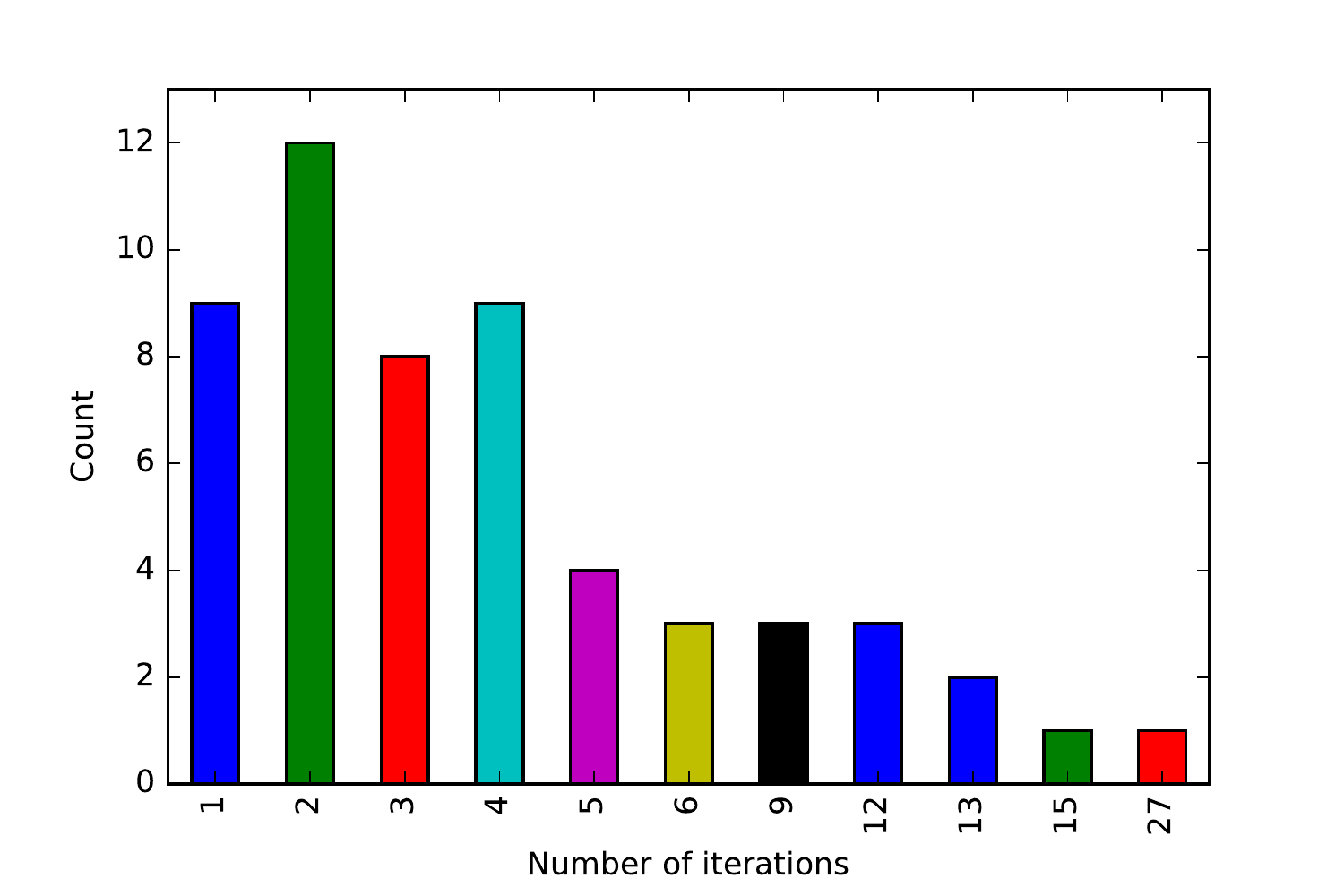}
	\caption{Histogram of the number of iterations per video frame when refining OSVOS results.}
    \label{fig:hist_iterations_other}
\end{figure}

It is not trivial to categorize user inputs into clicks, strokes or highlighting. 
Thus, rather than analyzing the number of annotated pixels, we use the number of refinement iterations to illustrate the amount of efforts the users invested. With a median of 4 iterative steps on both the first (Figure \ref{fig:hist_iterations_0}) and the refined frames (Figure \ref{fig:hist_iterations_other}) , we can observe that users tend to refine their masks only a few times.

After getting the first result, the user has the option to annotate more frames to refine the initial result. We find that only 15\% of our users invest time to do a refinement.
We can attribute this to our current user flow focused on annotating one frame.

\begin{figure}[t]
	\centering
	\includegraphics[width=0.7\linewidth]{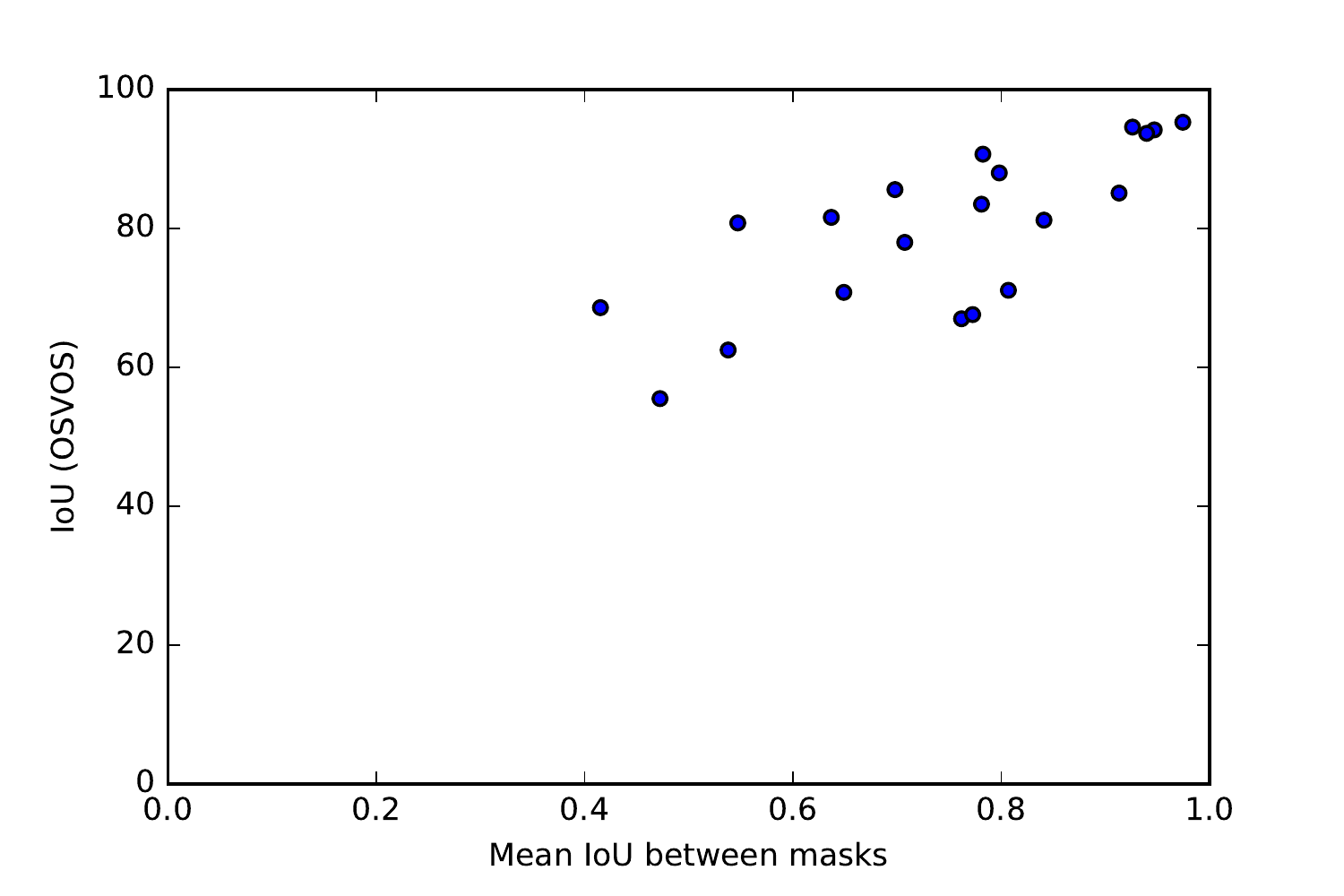}
	\caption{IOU overlap \vs OSVOS performance. The two are correlated, indicating that stronger frame to frame changes of the masks negatively affect OSVOS performance.}
    \label{fig:similar}
\end{figure}

\subsection{Future improvements}
From analyzing the user data, we find that people annotate inaccurately. While we currently treat user inputs as hard constraints, these inaccuracies suggest that the user input should be used as an indication only.

We find that our users mostly create stickers of people's faces and of pets such as cats and dogs. This indicates that pre-training OSVOS with a dataset of people and pets would improve the quality of a large portion of stickers.
We also did an analysis of the performance of OSVOS on the DAVIS dataset. Thereby, we noticed a correlation between the mean IoU between frame masks and OSVOS precision (Figure \ref{fig:similar}).
This shows which shows that fast moving objects are harder to segment. From a user experience point of view, it could make sense to predict the difficulty of a sequence due to motion or other factors (\eg using optical flow). Such a difficulty estimate would allow to guide users towards easier sequences and thus help them make better stickers.

\section{Conclusion}
In this paper, we have presented our method for fast and interactive video object segmentation.
Towards our goal of making video object segmentation practical in the wild, we have made several technical contributions. We have proposed a method to speed up the annotation of the first frame and have introduced a way for correcting mistakes of the video object segmentation, via a novel interactive segmentation model.
We have empirically evaluated the different components and the full system. Our experiments showed that our model is competitive or superior to existing methods for interactive segmentation. We also showed that using masks obtained with our interactive segmentation method, rather than perfect pixel-accurate masks, affects the OSVOS results only minimally.
Finally, we have provided insights into usage patterns. We have shown, for example, that most users tend to do very few iterations to annotate the first frame, thus highlighting the importance of a strong interactive segmentation model.
We hope that our work and the insights we provide will spur further research on making better video object segmentation tools.

{\small
\bibliographystyle{ieee}
\bibliography{references}
}

\end{document}